\documentclass{article} 
\usepackage{iclr2026_conference,times}

\usepackage{amsmath,amsfonts,bm}

\def\eqref#1{equation~\ref{#1}}

\def\1{\bm{1}}

\def\vc{{\bm{c}}}
\def\vd{{\bm{d}}}
\def\ve{{\bm{e}}}

\def\vp{{\bm{p}}}

\DeclareMathAlphabet{\mathsfit}{\encodingdefault}{\sfdefault}{m}{sl}
\SetMathAlphabet{\mathsfit}{bold}{\encodingdefault}{\sfdefault}{bx}{n}

\usepackage{hyperref}
\usepackage{url}
\usepackage{graphicx}
\usepackage{subcaption}
\usepackage{booktabs}

\newcommand{\Cam}{\mathcal{C}} 
\newcommand{\Lock}{\mathcal{L}} 
\newcommand{\Move}{\mathcal{M}} 
\newcommand{\Dodge}{\mathcal{D}} 
\newcommand{\HA}{\mathcal{H}} 

\newcommand{\piCam}{\pi_{\Cam}}
\newcommand{\piLock}{\pi_{\Lock}}
\newcommand{\piMove}{\pi_{\Move}}
\newcommand{\piDodge}{\pi_{\Dodge}}
\newcommand{\piHA}{\pi_{\HA}}

\title{Learning Transferable Skills in Action RPGs via Directed Skill Graphs and Selective Adaptation}

\author{Ali Najar \\
Sharif University of Technology\\
\texttt{anajar13750@gmail.com}
}

\iclrfinalcopy 

\begin{document}

\maketitle

\begin{abstract}
Lifelong agents should expand their competence over time without retraining from scratch or overwriting previously learned behaviors. We investigate this in a challenging real-time control setting (Dark Souls III) by representing combat as a directed skill graph and training its components in a hierarchical curriculum.
The resulting agent decomposes control into five reusable skills: camera control, target lock-on, movement, dodging, and a heal--attack decision policy, each optimized for a narrow responsibility.
This factorization improves sample efficiency by reducing the burden on any single policy and supports selective post-training: when the environment shifts from Phase 1 to Phase 2, only a subset of skills must be adapted, while upstream skills remain transferable.
 Empirically, we find that targeted fine-tuning of just two skills rapidly recovers performance under a limited interaction budget, suggesting that skill-graph curricula together with selective fine-tuning offer a practical pathway toward evolving, continually learning agents in complex real-time environments.
\end{abstract}


\section{Introduction}

Lifelong agents should continuously evolve their capabilities, integrating new skills without the inefficiency of retraining from scratch or the risk of overwriting established behaviors. This objective, often studied under continual and lifelong reinforcement learning, is challenging because the agent must balance plasticity (rapid adaptation) with stability (retaining useful skills) under non-stationarity and limited interaction budgets \citep{Hadsell2020EmbracingCC,Khetarpal2020TowardsCR}. In practice, monolithic end-to-end policies can be sample-inefficient and brittle when a task shifts, since the same parameters must simultaneously represent multiple competencies and their interactions.

These issues are amplified in complex real-time control domains such as modern video games. Most role-playing games feature tight reaction loops, partial observability, long-horizon credit assignment, and coupled subproblems. While large-scale deep RL has demonstrated that high-performance real-time game play is possible \citep{Vinyals2019GrandmasterLI, Berner2019Dota2W, novelRL2022Csereoka, oconnor2025learningdarksoulscombat}, achieving robust, transferable behavior typically requires extensive data and carefully engineered training protocols. A long-standing strategy for improving tractability is to introduce temporal abstraction and hierarchy~\citep{Pateria2021HRLSurvey}: the options framework formalizes temporally extended skills as reusable actions \citep{SUTTON1999BetweenMDP}, and the option-critic architecture learns such skills end-to-end \citep{Bacon2017OptionCritic}. More recently, multiple lines of work have emphasized {structuring} skills to improve exploration and composition, including graph-based skill representations \citep{Bagaria2021SkillDF,Lee2022DHRL}.

Complementary to hierarchy, modular architectures reduce interference and encourage reuse by decomposing control into reusable components. In multi-task and lifelong settings, modular policies can improve sample efficiency and transfer by reusing submodules rather than relearning entire policies per task \citep{Yang2020SoftModularization,MendezMendez2022ModularLR}. Curriculum learning offers a related lever: sequencing skills can substantially reduce the data needed in difficult domains \citep{Narvekar2020CurriculumLF}. Together, these ideas motivate a {directed skill graph} in which narrowly scoped skills are linked by explicit dependencies, providing a scaffold for efficient learning and selective adaptation.


In this work, this hypothesis is investigated in {Dark Souls III}, a challenging real-time combat environment. Recent work \citep{schuck_soulsgym_2023, towers2024gymnasium} has begun to establish {Dark Souls III} as an RL research platform, often via memory instrumentation that exposes compact state signals for training. We model combat control as a directed skill graph whose nodes are five reusable skills: camera control, target lock-on, movement, dodging, and a heal--attack decision policy. The components are trained in a hierarchical curriculum aligned with the dependency structure. Skills are trained sequentially for sample efficiency but executed concurrently at run time, enabling upstream skills to remain reusable while downstream skills specialize.

A core motivation in lifelong learning is {selective post-training} under domain shift: when the environment changes, only the most sensitive components should be adapted while others remain stable and transferable. This aligns with rapid adaptation via fine-tuning \citep{Finn2017ModelAgnosticMF} and with continual RL methods emphasizing forward transfer and reuse \citep{Woczyk2022DisentanglingTI, Wang2020LifelongIR}. Selective adaptation is evaluated by splitting a boss encounter into Phase~1 and Phase~2 domains and measuring both zero-shot transfer and targeted fine-tuning. Empirically, downstream ablations sharply degrade performance while upstream skills retain utility across domains, and fine-tuning a small subset of skills rapidly recovers performance under a limited interaction budget.


\paragraph{Contributions.}
(i) We formulate Dark Souls III combat as a directed skill graph and instantiate a modular agent with five reusable skills.
(ii) We propose a hierarchical training protocol that improves sample efficiency by isolating narrow competencies and reusing previously learned skills.
(iii) We demonstrate selective post-training under a Phase~1 to Phase~2 domain shift, supported by ablations that characterize which skills transfer and which require adaptation.

\section{Skill Graph and Hierarchical Training}
\label{sec:method}

\begin{figure}[t]
  \centering
    \includegraphics[width=\linewidth]{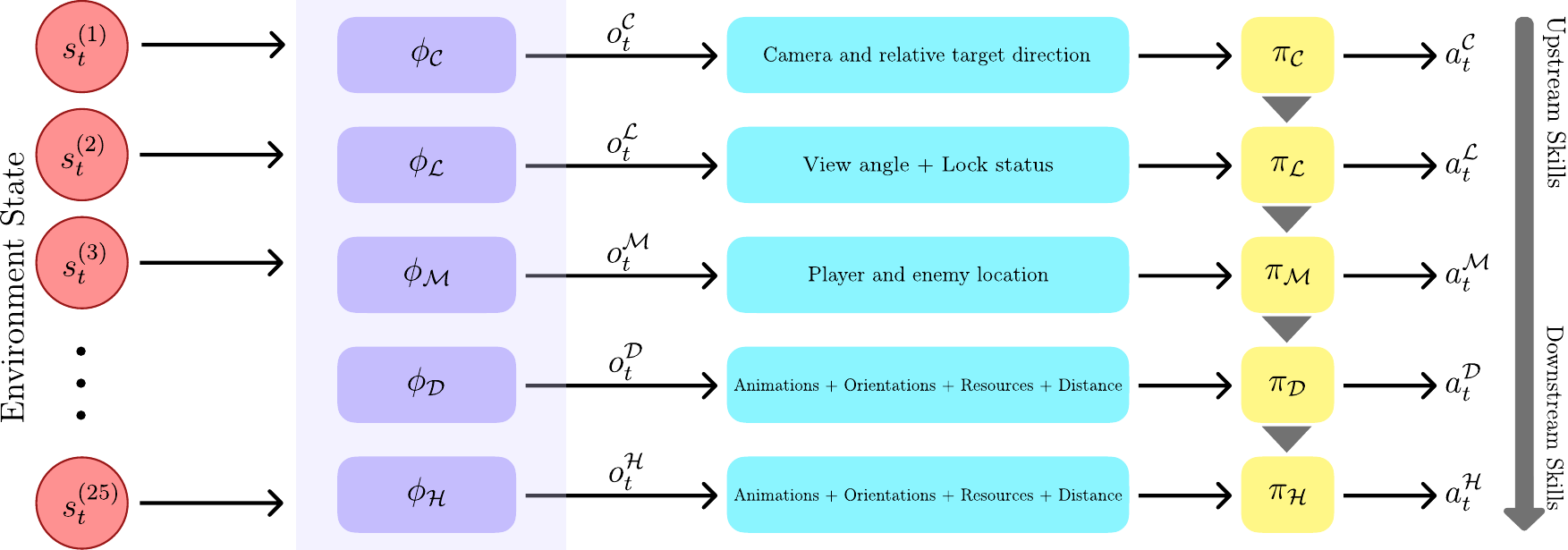}
  \caption{\textbf{The Modular Skill Graph Architecture.} At run-time, the global environment state (with dimension 25 in our experiments) is mapped into five specialized observation spaces ($o_t^k$), which are processed concurrently by independent skill policies ($\pi_k$) to generate action components. The grey arrows between policies denote the hierarchical training dependency: upstream skills (top) are trained first and held fixed, shaping the data distribution for the adaptive downstream skills (bottom).}
  \label{fig:skill_graph}
\end{figure}

Our design targets two constraints typical of real-time game control: (i) decision-making comprises heterogeneous subproblems (view control, targeting, positioning, defense, and resource/action selection), and (ii) post-training updates should be localized under domain shift. Accordingly, a modular controller is used in which each component has a narrow responsibility and a small action set, reducing entanglement and enabling selective adaptation.


\paragraph{Skill Set.}
We implement five skills, denoted by
$\Cam$ (camera control), $\Lock$ (lock-on), $\Move$ (movement/positioning), $\Dodge$ (dodging), and $\HA$ (heal--attack decisions),
with corresponding policies $\piCam, \piLock, \piMove, \piDodge,$ and $\piHA$.
Each skill $k \in \{\Cam,\Lock,\Move,\Dodge,\HA\}$ receives its own observation
$o_t^{k} = \phi_{k}(s_t)$, where $s_t$ is the available game state and $\phi_{k}$ selects variables relevant to that responsibility (Figure~\ref{fig:skill_graph}).
For completeness, the full specification of each skill's observation features is provided in Appendix~\ref{app:environment}.

\paragraph{Policy Composition.}
At each control step, every skill policy $\pi_k$ produces a low-dimensional control output $a_t^{k}$.
This mirrors how humans control the game: view control, positioning, and combat decisions are issued concurrently rather than as a single atomic action.
We execute the agent by composing these per-skill outputs into a single control signal,
\begin{equation}
a_t \;=\; \mathcal{C}\big(a_t^{\Cam}, a_t^{\Lock}, a_t^{\Move}, a_t^{\Dodge}, a_t^{\HA}\big),
\label{eq:compose}
\end{equation}
where $\mathcal{C}(\cdot)$ is a fixed composition operator.
This composition is mostly straightforward because most of the skills occupy a distinct control subspace.
In our implementation, skills are evaluated concurrently (multi-threaded) and the control loop applies $a_t$ at a fixed rate by merging the {most recent} outputs from each skill.
This yields a practical approximation to synchronous composition.
More details of $\mathcal{C}$ and the per-skill action spaces are provided in Appendix~\ref{app:environment}.

\subsection{Curriculum Learning}
\label{subsec:curriculum}
Figure~\ref{fig:skill_graph} summarizes the training dependency chain,
\begin{equation}
\Cam \rightarrow \Lock \rightarrow \Move \rightarrow \Dodge \rightarrow \HA,
\label{eq:skillgraph}
\end{equation}
where $i \rightarrow j$ means $\pi_j$ is trained with all upstream policies fixed.
Training proceeds sequentially from left to right.
When optimizing $\pi_k$, we load and freeze $\{\pi_j: j \prec k\}$, and train only $\pi_k$ under the closed-loop behavior induced by the frozen upstream skills.
This staging reduces effective exploration burden for later skills by constraining the reachable state distribution to task-relevant configurations.

\paragraph{Downstream Adaptation.}
Because skills are trained in a staged curriculum, each downstream policy is optimized in the presence of fixed upstream behavior and therefore must adapt to it.
Upstream skills also reshape the downstream learning problem by constraining reachable states and shifting the visitation distribution.
For instance, a competent $\Cam$ and $\Lock$ stabilize viewpoint and targeting, and $\Move$ biases encounters toward meaningful engagement configurations, which reduces the exploration burden for $\Dodge$ and $\HA$ and typically accelerates training.

\paragraph{Cooperation.}
Upstream skills define constraints that downstream skills should respect.
For example, if $\HA$ is trained or enabled before $\Dodge$ is reliable, an aggressive $\HA$ policy may over-commit to attacks and incur frequent hits; conversely, once $\Dodge$ is competent, $\HA$ should learn to avoid actions that systematically undermine defensive timing.
Thus, the curriculum encourages \textit{cooperative specialization}:  each skill optimizes its objective while minimizing interference with established upstream competencies.

\paragraph{Per-skill Objectives.}
Each skill is trained with its own reward designed to reflect its narrow responsibility and to avoid encoding boss-specific scripts.
At a high level, $\Cam$ is rewarded for improving viewpoint alignment to the target, $\Lock$ for maintaining a valid lock state, $\Move$ for encouraging purposeful positioning and engagement rather than wandering, $\Dodge$ for avoiding damage while respecting resource constraints, and $\HA$ for trading off damage dealt, damage taken, and heal usage.
The exact reward definitions used in our experiments are reported in Appendix~\ref{app:reward}.

\paragraph{Selective Fine-tuning Under Phase Shift.}
A key motivation for the skill graph is localized post-training when the domain changes.
We do this by transferring from the boss encounter in Phase~1 to Phase~2.
Because upstream skills primarily capture phase-invariant mechanisms ($\Cam, \Lock, \Move$), we keep $\pi_{\Cam}$, $\pi_{\Lock}$, and $\pi_{\Move}$ fixed during Phase~2 adaptation, and fine-tune only the phase-sensitive skills ($\pi_{\Dodge}$ and $\pi_{\HA}$).
Section~\ref{sec:experiments} reports both zero-shot transfer and the effect of targeted fine-tuning under a limited interaction budget.

\paragraph{Skill Ablations.}
To characterize specialization and dependencies, we perform controlled ablations by replacing a selected policy $\pi_k$ with a random policy over its action space while keeping all other skills unchanged.
This provides a direct test of whether a skill is necessary for the composed agent's performance and whether upstream skills remain useful when downstream skills are removed. The results are reported in Section~\ref{sec:experiments}.

\section{Experiments}
\label{sec:experiments}

\begin{figure}[t]
  \centering
  \begin{subfigure}[b]{0.32\linewidth}
    \centering
    \includegraphics[width=\linewidth]{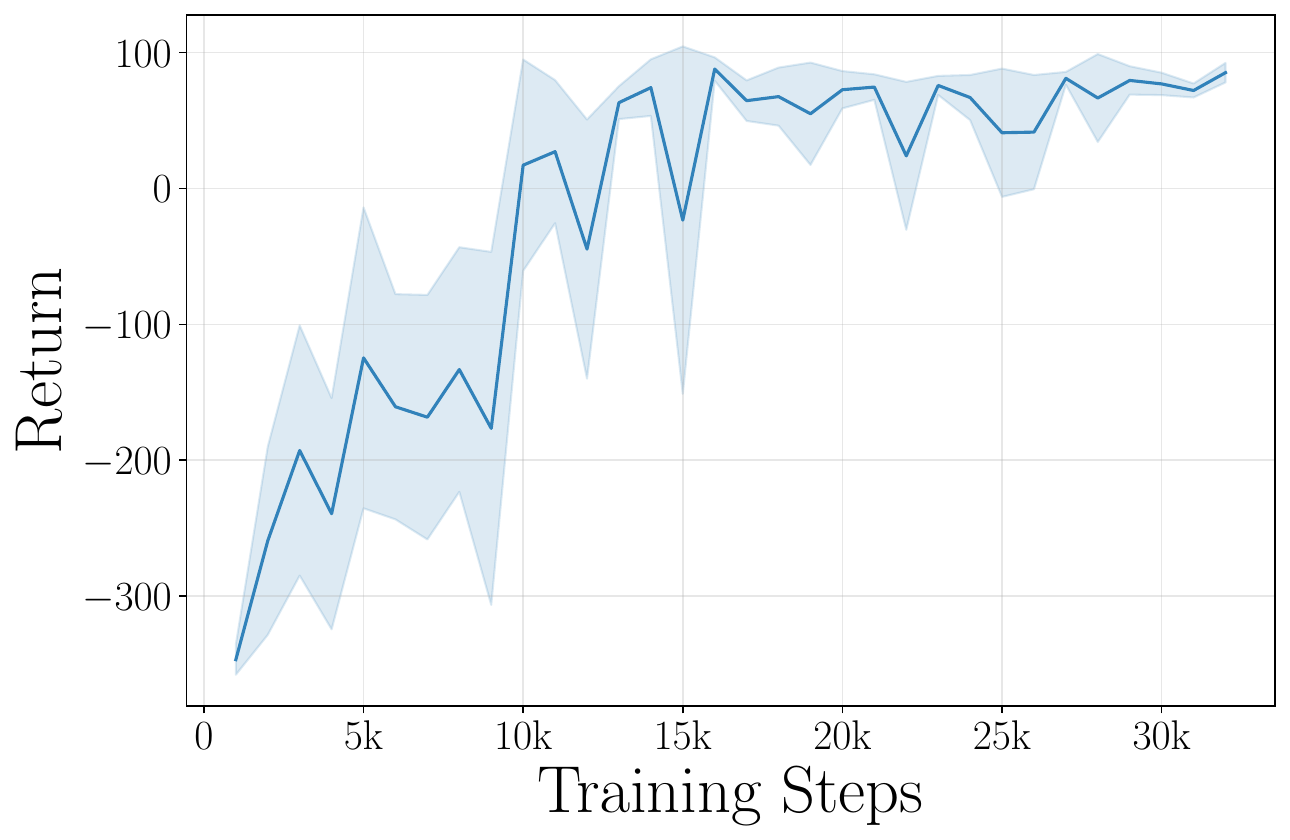}
    \caption{Camera}
    \label{fig:cam}
  \end{subfigure}\hfill
  \begin{subfigure}[b]{0.32\linewidth}
    \centering
    \includegraphics[width=\linewidth]{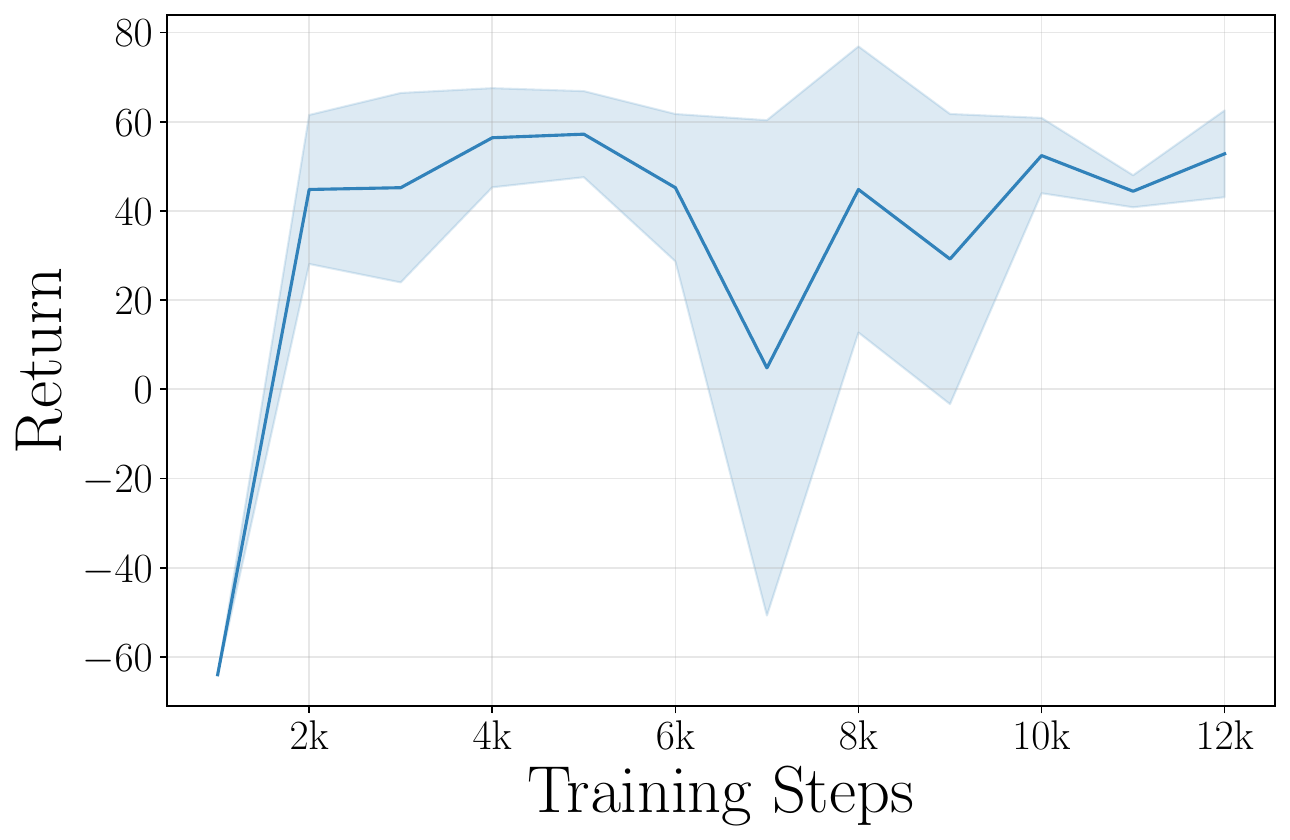}
    \caption{Lock-on}
    \label{fig:lock}
  \end{subfigure}\hfill
  \begin{subfigure}[b]{0.32\linewidth}
    \centering
    \includegraphics[width=\linewidth]{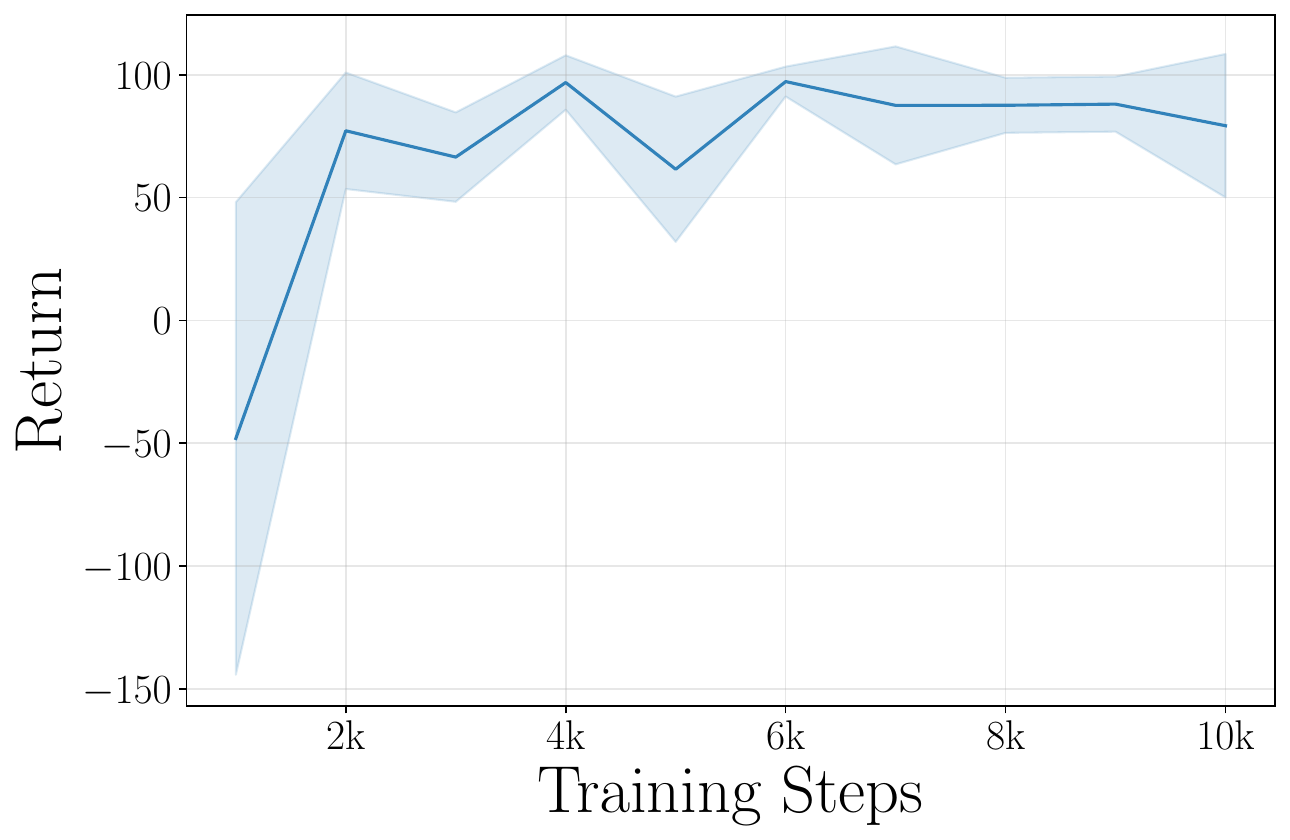}
    \caption{Movement}
    \label{fig:move}
  \end{subfigure}

  \caption{Return vs.\ training steps for $\Cam$, $\Lock$, and $\Move$, averaged over five evaluation episodes at each 1k-step checkpoint. The shaded region denotes the 95\% confidence interval.}
  \label{fig:upstream}
\end{figure}

\begin{figure}[t]
  \centering
  \begin{subfigure}[b]{0.49\linewidth}
    \centering
    \includegraphics[width=\linewidth]{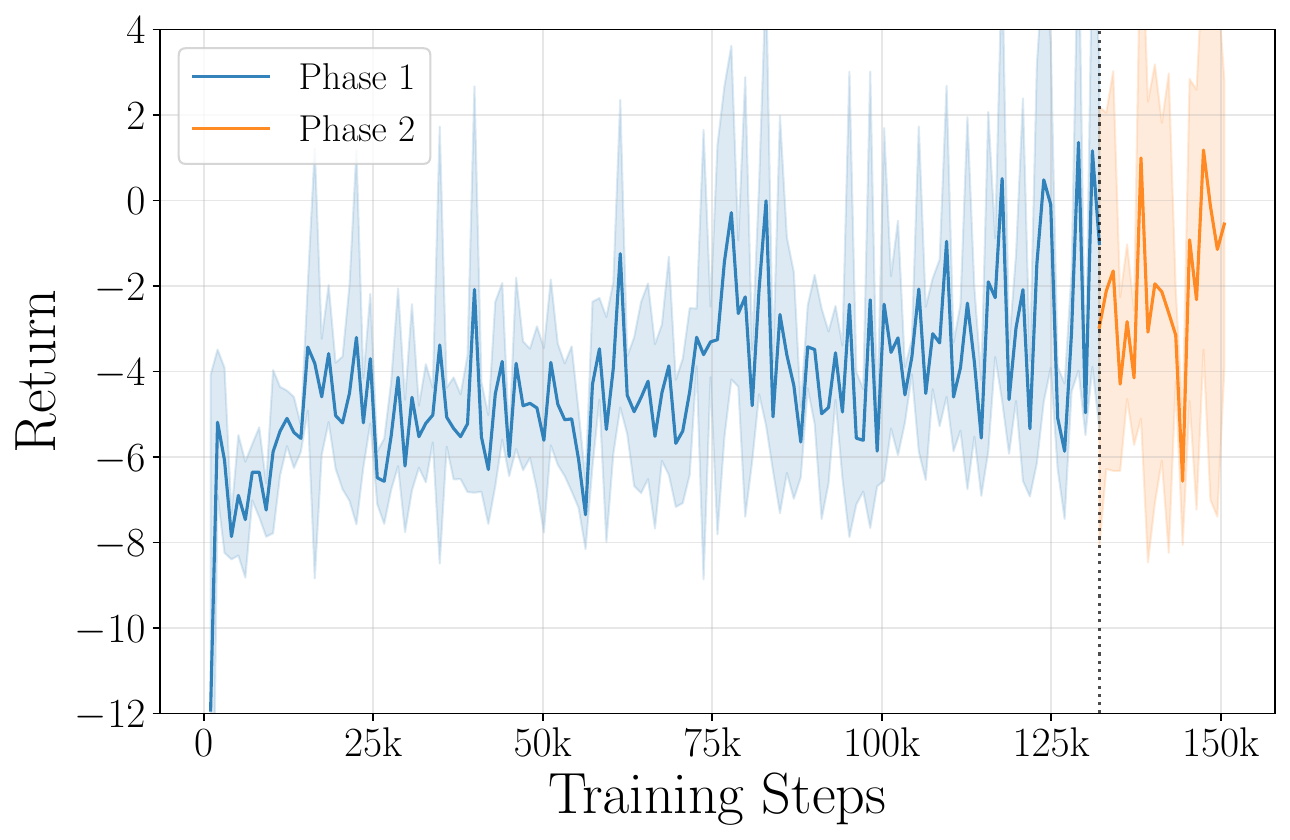}
    \caption{Dodge}
    \label{fig:dodge}
  \end{subfigure}\hfill
  \begin{subfigure}[b]{0.49\linewidth}
    \centering
    \includegraphics[width=\linewidth]{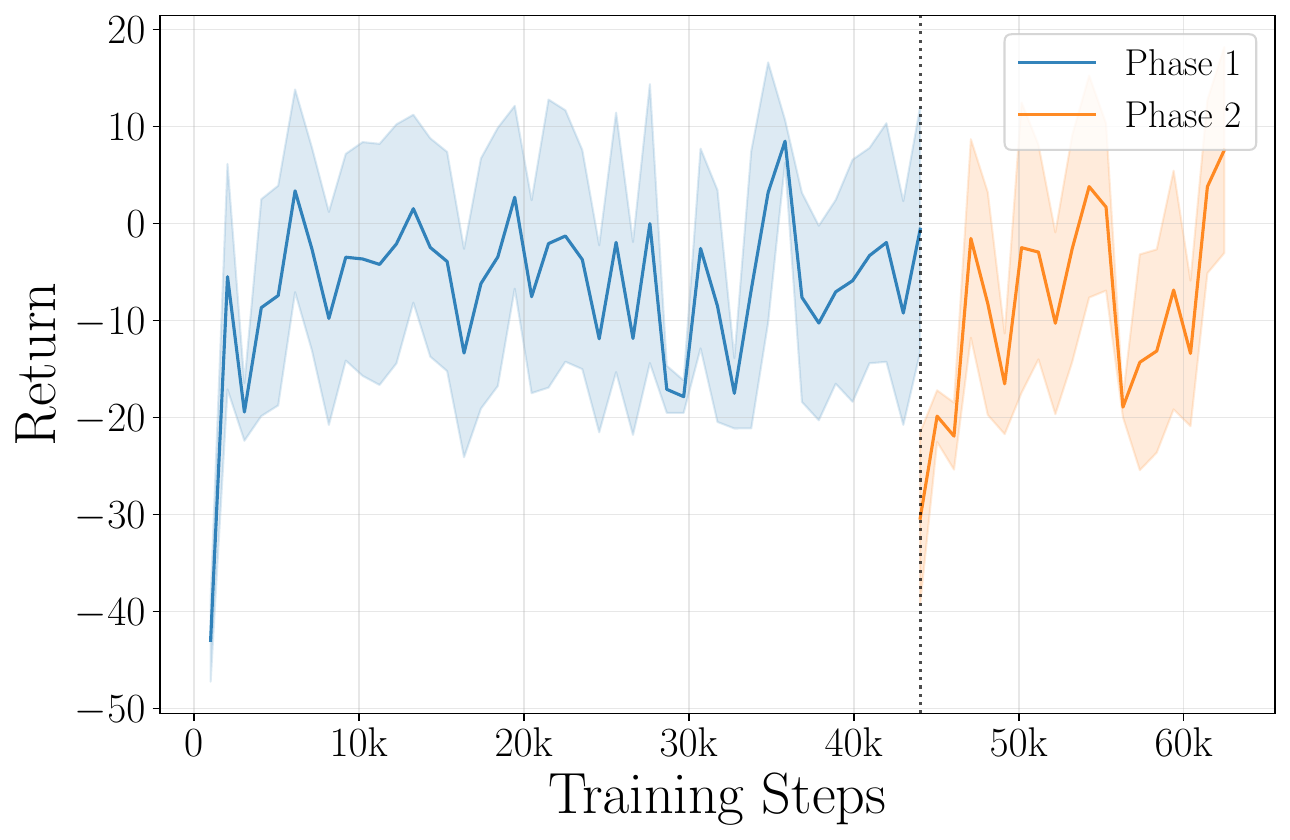}
    \caption{Heal--attack}
    \label{fig:ha}
  \end{subfigure}

  \caption{Return vs.\ training steps for $\Dodge$ and $\HA$, averaged over five evaluation episodes at each 1k-step checkpoint. The shaded region denotes the 95\% confidence interval, and the dashed vertical line marks the Phase~1 $\rightarrow$ Phase~2 transition.}
  \label{fig:downstream}
\end{figure}

\subsection{Setup}
\label{subsec:exp_setup}

\paragraph{Game Interface and Observations.}
A process-memory state readout interface is used for {Dark Souls III}.
Cheat Engine~\citep{cheatengine_2025} is used to locate the required memory variables (e.g., position, pose, resources, lock status, and animation signals).
Pixel-based perception is not evaluated in this work due to compute and engineering constraints; instead, this compact state interface is used to keep the focus on skill decomposition, transfer, and selective fine-tuning.
Full details of the experiments are provided in Appendix~\ref{app:experimental_details}.


\paragraph{Learning Algorithm.}
We intentionally use a simple, widely-used value-based baseline, Deep Q-Networks (DQN) \citep{mnih2013playingatarideepreinforcement}, for all skills.
The goal is not to maximize raw performance through algorithmic sophistication, but to test whether the {skill-graph factorization} and {selective fine-tuning} are sufficient to produce transferable behavior.
This is a conservative choice: vanilla DQN is known to be sensitive to exploration design, brittle under non-stationarity, and prone to interference when used in continual settings (i.e., it provides no explicit mechanism to prevent forgetting) \citep{Pateria2021HRLSurvey, Khetarpal2020TowardsCR}.  

\subsection{Results}
\label{subsec:exp_results}

\paragraph{Sample Efficiency.}
Figure~\ref{fig:upstream} shows that the upstream skills ($\Cam$, $\Lock$, $\Move$) are learned quickly under the proposed curriculum, reaching near-maximal return.
To contextualize sample efficiency, a single atomic end-to-end agent is trained on the same Phase~1 encounter using the same state interface and a comparable DQN-based setup.
With the proposed skill graph, a competitive Phase~1 policy is obtained with an overall interaction budget of approximately 230k steps (Table~\ref{tab:winrates_p1}). In contrast, the atomic end-to-end baseline gets no where close to learning a reliable combat behavior even after plenty of steps (Table~\ref{tab:winrates_p1}).
The return curve for the end-to-end baseline over training steps and further discussions are reported in Appendix~\ref{app:end2endbase}.

\paragraph{Downstream Skills.}
The downstream skills are more challenging and account for the majority of interaction.
$\Dodge$ requires precise timing and is the hardest component to learn.
This difficulty is reflected in Figure~\ref{fig:dodge} as poorly timed dodges can be harmful, making early exploration misleading and slow to improve. The gradual increase in return indicates that the agent is learning to survive longer and reduce early deaths.
For interpretation, note that the maximum achievable return for $\pi_\Dodge$ under our shaping is approximately 10, while returns near 0 correspond to ``survival for a non-trivial duration'' (i.e., not dying quickly), which already indicates meaningful defensive competence.

\paragraph{$\HA$ Dependence on $\Dodge$.}
The $\HA$ policy typically reaches a reasonable attack strategy quickly, but reliable healing is hindered due to structural data sparsity: the number of healing opportunities is capped by the environment, making credit assignment particularly challenging for DQN.
Under our reward scale, the maximum return for $\pi_\HA$ is approximately 15. Figure~\ref{fig:ha}  shows that the agent wins in fewer than half of the trials. One more thing to note is that if $\pi_\Dodge$ is imperfect, $\pi_\HA$ is incentivized to behave more aggressively, attempting to end the fight before accumulating fatal damage, which can reduce the learning signal for conservative healing.

\paragraph{Ablations Confirm Specialization.}
Table~\ref{tab:winrates} shows that downstream skills are critical for overall success.
With both $\pi_\Dodge$ and $\pi_\HA$ randomized, performance drops to 0\%.
Randomizing only $\pi_\Dodge$ reduces win rate to 16\%, while randomizing only $\pi_\HA$ reduces it to 4\%.
The higher win rate under a randomized $\pi_\Dodge$ reflects a shift toward aggression, where ending the fight quickly is the only viable path when defense is unreliable.
With both trained, the composed policy achieves a 44\% win rate in Phase~1, indicating that wins depend on competent downstream timing and decision-making.

\paragraph{Transfer.}
When transferring the Phase~1-trained system to Phase~2 without additional training, non-trivial zero-shot performance is obtained (33.3\% mid-range starts; 12.5\% long-range starts; Table~\ref{tab:winrates_p2}).
Because the Phase~2 boss has higher health and damage but is less aggressive at close range, transfer is evaluated under different initializations: episodes are started both at mid-range and at longer range from the boss to probe how well the Phase~1 $\pi_\Dodge$ behavior generalizes across engagement distances.
Selective fine-tuning of only $\pi_\Dodge$ and $\pi_\HA$ (mid-range starts) improves the Phase~2 win rate to 52\%, demonstrating that adaptation can be localized to a small subset of policies under a limited interaction budget.

\begin{table}[t!]
\centering
\caption{Win rates (\%) measured over 25 episodes per setting. ``Random'' replaces the corresponding policy with a uniform random policy over its action space, while keeping other skills fixed.}

\label{tab:winrates}

\footnotesize
\setlength{\tabcolsep}{3.5pt}
\renewcommand{\arraystretch}{1.05}

\begin{subtable}[t]{0.49\linewidth}
\centering
\caption{Phase~1}
\label{tab:winrates_p1}
\begin{tabular}{l c}
\toprule
\textbf{Setting} & \textbf{Win (\%)} \\
\midrule
$\pi_\HA$ random, $\pi_\Dodge$ random & 0.0 \\
$\pi_\HA$ random, $\pi_\Dodge$ trained & 4.0 \\
$\pi_\HA$ trained, $\pi_\Dodge$ random & 16.0 \\
$\pi_\HA$ trained, $\pi_\Dodge$ trained & \textbf{44.0} \\
Single atomic agent (end-to-end) & 0.0 \\
\bottomrule
\end{tabular}
\end{subtable}\hfill
\begin{subtable}[t]{0.49\linewidth}
\centering
\caption{Phase~2}
\label{tab:winrates_p2}
\begin{tabular}{l c}
\toprule
\textbf{Setting} & \textbf{Win (\%)} \\
\midrule
Zero-shot transfer (mid-range start) & 33.3 \\
Zero-shot transfer (long-range start) & 12.5 \\
Selective fine-tuning ($\pi_\Dodge + \pi_\HA$ only) & \textbf{52.0} \\
\bottomrule
\end{tabular}
\end{subtable}
\end{table}

\section{Conclusion}
We studied lifelong adaptation in a challenging real-time domain ({Dark Souls III}) by modeling it as a directed skill graph trained via a hierarchical curriculum. We achieved sample-efficient learning far exceeding a monolithic baseline. Crucially, this factorization enabled selective post-training: when the domain shifted, performance was recovered by fine-tuning only downstream, phase-sensitive skills under a limited interaction budget while reusing upstream behaviors. These results suggest that structuring agents around skill dependencies is a promising pathway for scalable lifelong learning in complex environments.

\bibliography{iclr2026_conference}
\bibliographystyle{iclr2026_conference}

\clearpage

\appendix

\section{Environments}
\label{app:environment}

\subsection{Dark Souls III Encounter and Phase Split}
\label{app:encounter}

The first boss encounter in {Dark Souls III}, \textit{Iudex Gundyr}, was used as the evaluation environment.
The boss has a total health pool of $1037$ HP, and the fight was partitioned into two phases to be analyzed as distinct domains.
Phase~1 was defined over approximately $415$ HP, while Phase~2 was defined over approximately $500$ HP, reflecting the change in the boss move set and combat dynamics. A typical player light attack deals roughly $30$--$50$ damage per hit.

To keep the boss loaded in the arena throughout training (avoiding the in-game defeat sequence and expensive reload/reset logic), episodes are terminated before the boss reaches zero HP.
Concretely, in Phase~1 we end an episode once the boss health falls below $622$.
In Phase~2 we end an episode once the boss HP falls below $60$.
These thresholds ensure the encounter remains persistent while still providing a consistent success signal for learning and evaluation.
Across both phases.

\begin{figure}[h!]
  \centering
  \begin{subfigure}[b]{0.49\linewidth}
    \centering
    \includegraphics[width=0.495\linewidth]{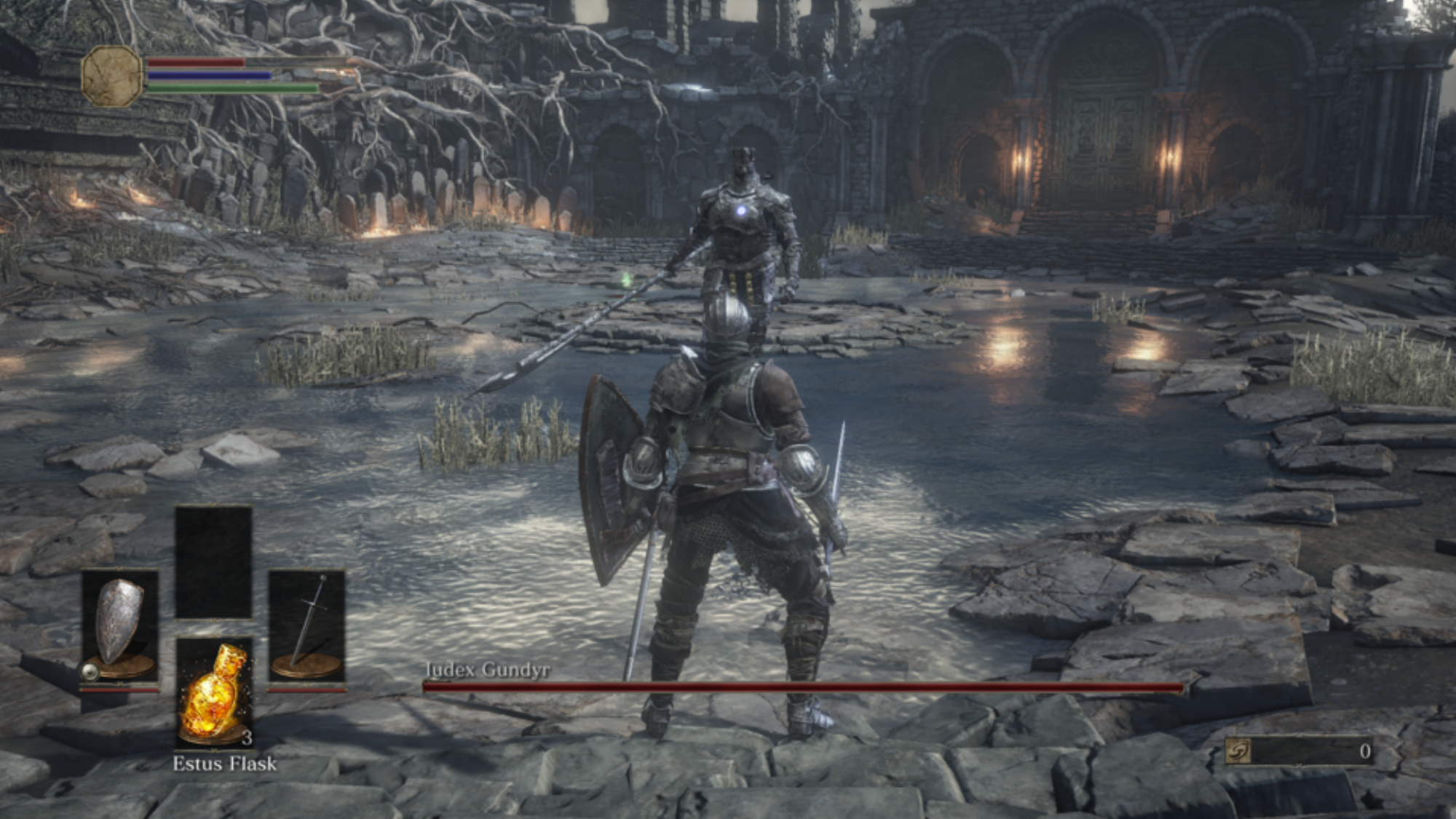}%
    \includegraphics[width=0.495\linewidth]{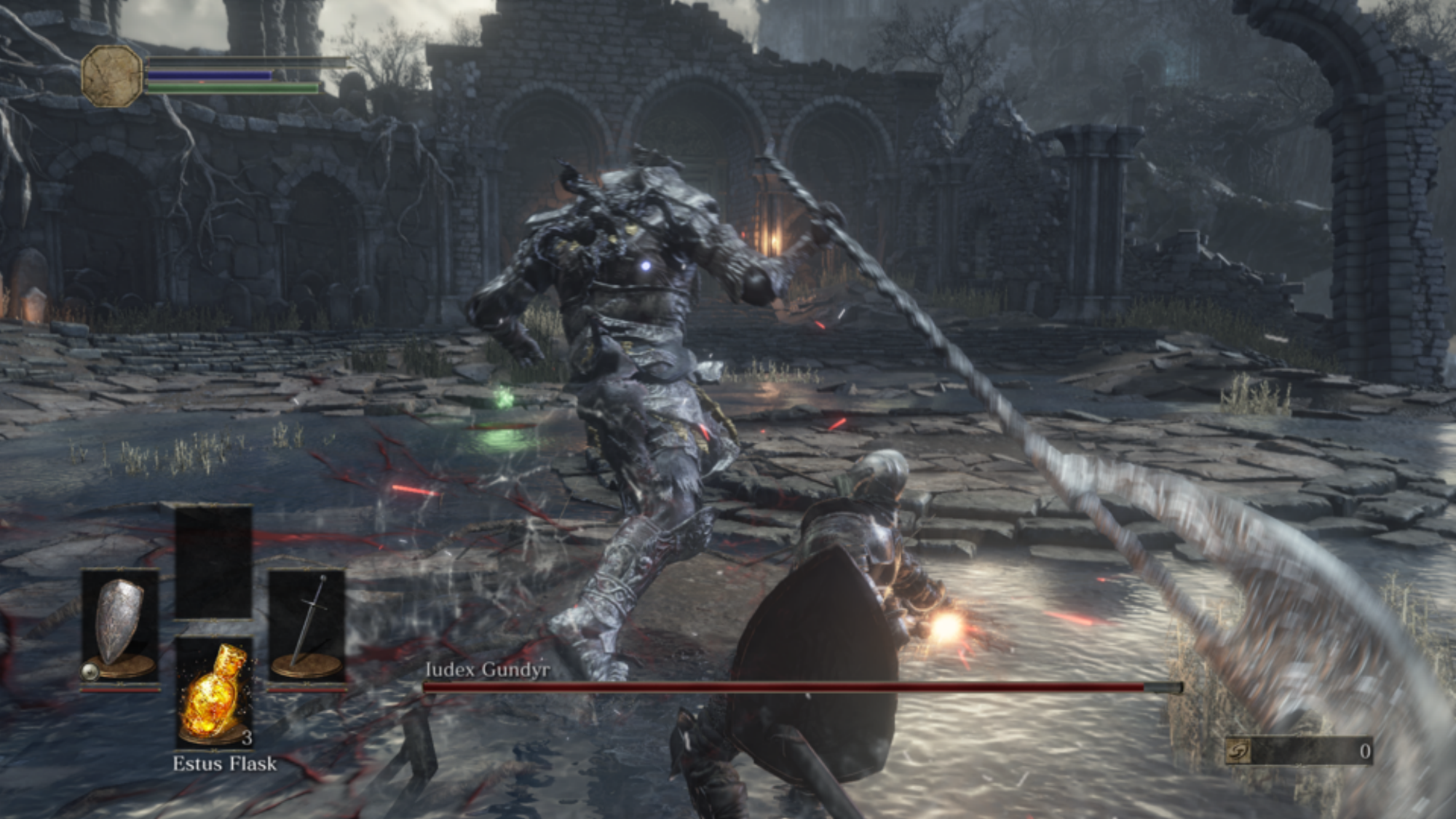}
    \caption{Phase~1.}
    \label{fig:iudex_p1}
  \end{subfigure}\hfill
  \begin{subfigure}[b]{0.49\linewidth}
    \centering
    \includegraphics[width=0.495\linewidth]{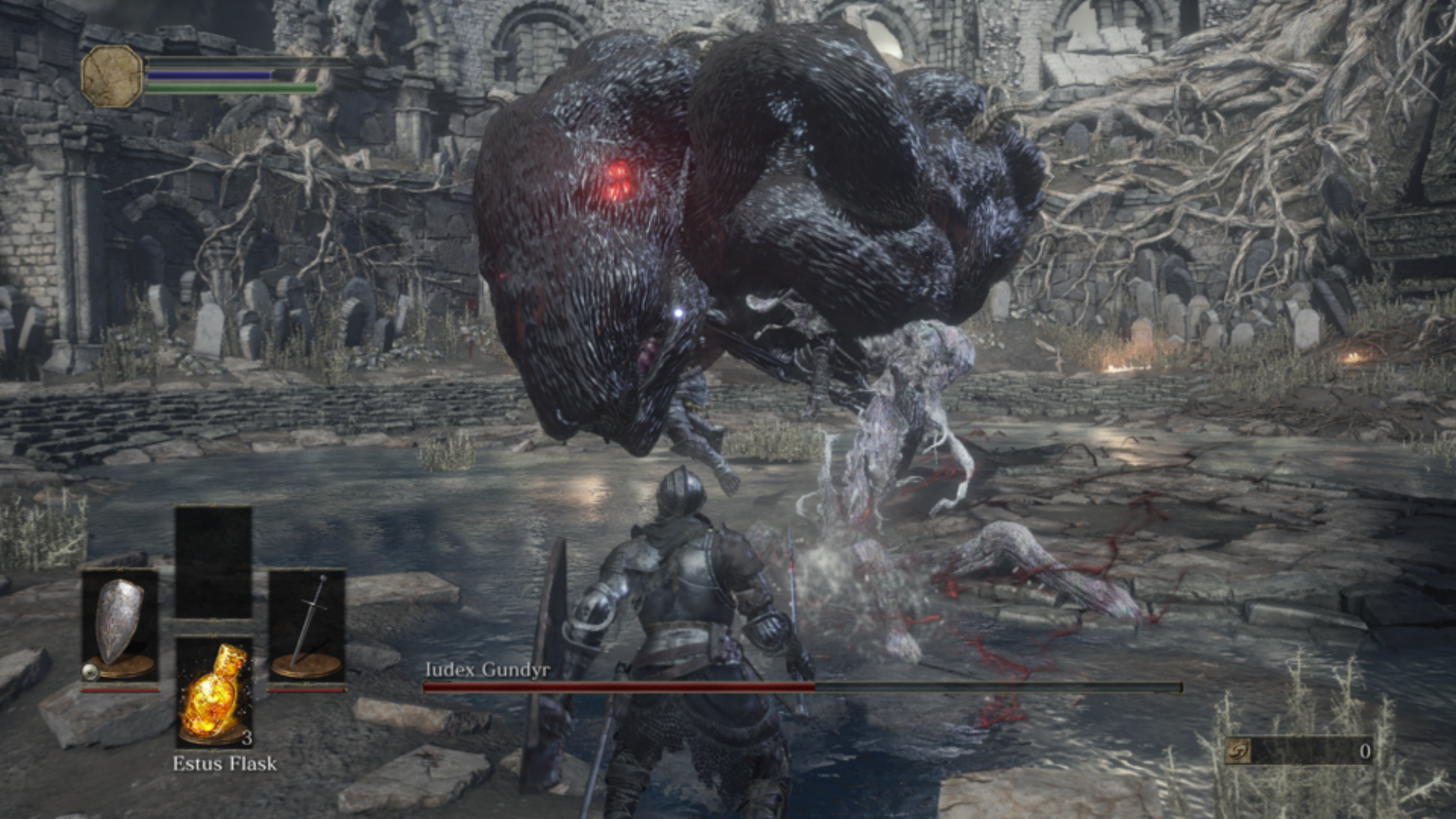}%
    \includegraphics[width=0.495\linewidth]{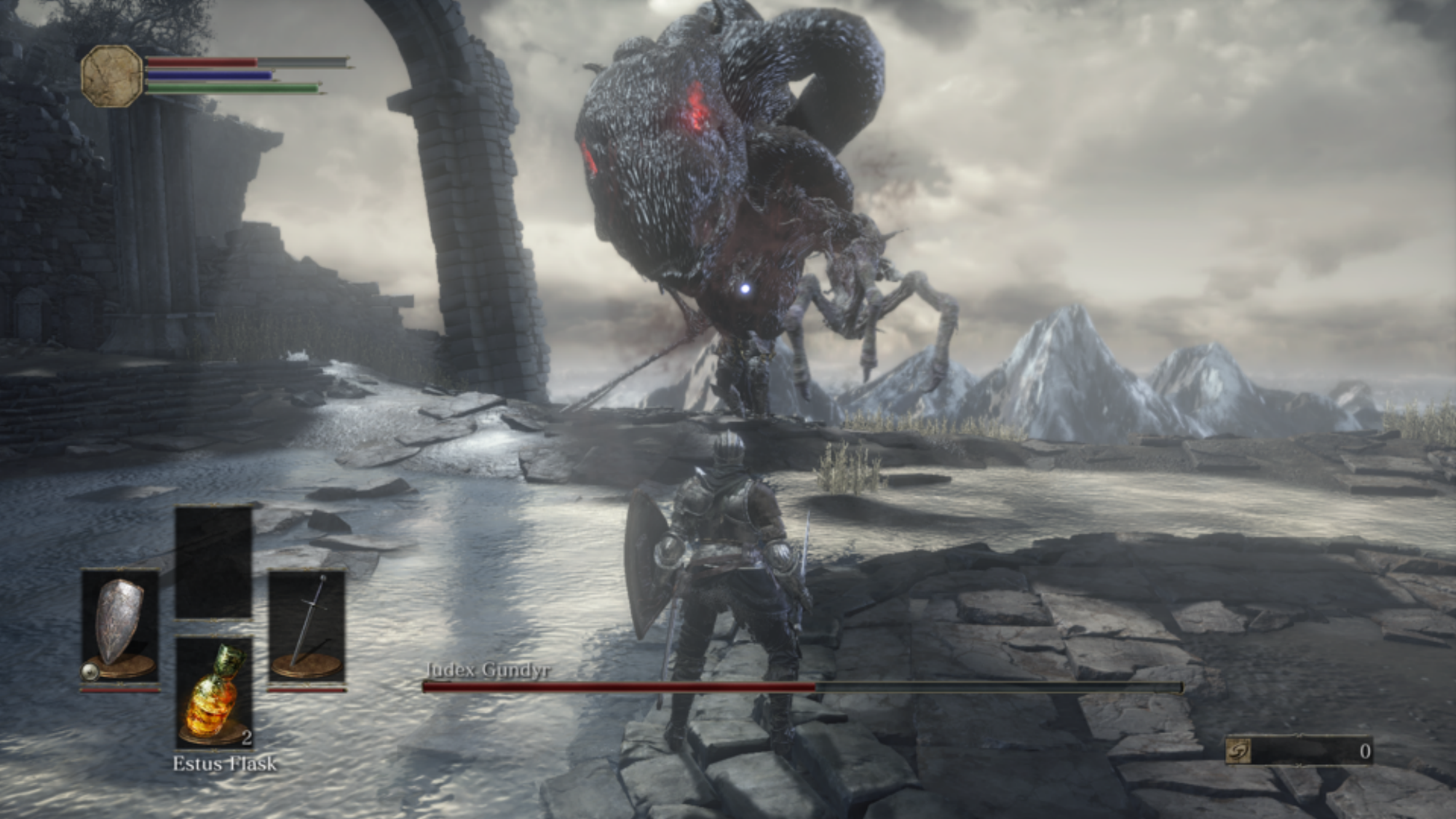}
    \caption{Phase~2.}
    \label{fig:iudex_p2}
  \end{subfigure}
  \caption{\textbf{Phase split used as a domain shift.} Each subfigure shows a concatenation of two screenshots from the corresponding phase. The encounter was divided into Phase~1 and Phase~2 to evaluate transfer and selective fine-tuning under a controlled shift in boss behavior.}
  \label{fig:iudex_phases}
\end{figure}

To avoid reward and value-scale issues, health variables provided to the RL algorithm were normalized by their corresponding maxima.
In particular, player and boss health were represented as ratios in $[0,1]$, which kept reward magnitudes comparable across phases and reduced sensitivity to absolute HP scaling. A base-level character instantiated with the default \textit{Knight} class was used for all runs.
Figure~\ref{fig:iudex_phases} illustrates the visual appearance of the two domains used in our experiments.

Each episode begins with the player positioned in the boss arena and ends on either boss defeat, player death, or a fixed time limit (skill-dependent in Appendix~\ref{app:experimental_details}).

\subsection{Observation and Action Spaces}
\label{app:obs_and_action}

\paragraph{Process-Memory Interface.}
A compact state interface is obtained via process-memory readout.
Cheat Engine~\citep{cheatengine_2025} is used offline to locate the relevant variables and resolve stable {relative} addresses.
During data collection and training, these variables are read from memory using the Python interface (\texttt{pyMeow}); Cheat Engine is not used in the training loop.
The resulting interface exposes positions, orientations, lock status, resources, and animation signals required by the skill policies.

\paragraph{Global State and Derived Features.}
Let the raw memory variables at time $t$ be grouped into a global state $s_t$ (dimension $25$ in our experiments; Figure~\ref{fig:skill_graph}).
From $s_t$, skill-specific observations are formed as $o_t^k=\phi_k(s_t)$ via lightweight feature construction.
The following derived quantities are used repeatedly:
\begin{align}
\vp_t &= (x_t,y_t,z_t) \in \mathbb{R}^3 \quad\text{(player position)}, \\
\ve_t &= (x^e_t,y^e_t,z^e_t) \in \mathbb{R}^3 \quad\text{(enemy position)}, \\
\vd_t &= \ve_t-\vp_t, \\
d_t &= \|\vd_t\|_2 \quad\text{(distance)}, \\
\hat{\vd}_t &= \vd_t / \|\vd_t\|_2 \in \mathbb{R}^3 \quad\text{(unit direction to enemy)}, \\
\vc_t &= (c^x_t,c^y_t,c^z_t) \in [-1, 1]^3 \quad\text{(camera direction vector)}, \\
\alpha_t &= \arccos\!\left(\frac{\langle \vc_t,\hat{\vd}_t\rangle}{\|\vc_t\|_2}\right) \in [0,\pi] \quad\text{(camera--target angle)}. \label{eq:alpha}
\end{align}

In Equation~\eqref{eq:alpha}, $\alpha_t \in [0,\pi]$ is the angle between the camera direction $\vc_t$ and the normalized enemy direction $\hat{\vd}_t$.
Enemy and player orientations are read as angles in radians, $O^e_t, O^p_t \in [-\pi,\pi]$.
Health and stamina are normalized to $[0,1]$, denoted by $h_t$ (player HP ratio), $h^e_t$ (enemy HP ratio), and $\sigma_t$ (stamina ratio). We also show the target lock-on status with $\ell_t \in \{0,1\}$.
Animation state is represented by an integer animation identifier $\kappa_t$ and a normalized progress ratio $\rho_t \in [0,1]$, computed as
$\rho_t = \tfrac{\texttt{anim\_time}}{\texttt{max\_anim\_time}}$ for the currently active animation.
Additionally, the remaining number of healing Estus flasks available in the episode is denoted by $n^{\text{estus}}_t\in \mathbb{N}$.

\begin{table}[h!]
\centering
\caption{Skill-specific observation vectors used for training each of the agents. Each skill receives $o_t^k=\phi_k(s_t)$.}
\label{tab:obs_spaces}
\footnotesize
\setlength{\tabcolsep}{5.0pt}
\renewcommand{\arraystretch}{1.15}
\begin{tabular}{l c c}
\toprule
\textbf{Skill} & \textbf{Observation $o_t^k$} & \textbf{Dim.} \\
\midrule
$\Cam$  & $[\hat{\vd}_t^{\top},\ \vc_t^{\top},\ \alpha_t]$ & 7 \\
$\Lock$ & $[\alpha_t,\ \ell_t]$ & 2 \\
$\Move$ & $[\vp_t^{\top},\ \ve_t^{\top}]$ & 6 \\
$\Dodge$ & $[\kappa^e_t,\ \rho^e_t,\ O^e_t,\ O^p_t,\ \sigma_t,\ h_t,\ d_t]$ & 7 \\
$\HA$ &
$[\kappa^e_t,\ \rho^e_t,\ \kappa^p_t,\ \rho^p_t,\ \sigma_t,\ h_t,\ h^e_t,\ O^e_t,\ O^p_t,\ n^{\text{estus}}_t,\ d_t]$
& 11 \\
\bottomrule
\end{tabular}
\end{table}

\begin{table}[h!]
\centering
\caption{Skill action spaces used for each of the agents. Actions are discrete and are applied using the keybindings of the game.}
\label{tab:action_spaces}
\footnotesize
\setlength{\tabcolsep}{6.0pt}
\renewcommand{\arraystretch}{1.15}
\begin{tabular}{l c c}
\toprule
\textbf{Skill} & \textbf{Action set $\mathcal{A}_k$} & \textbf{Dim.} \\
\midrule
$\Cam$  & \{cam up, cam down, cam left, cam right, idle\} & 5 \\
$\Lock$ & \{toggle lock-on, idle\} & 2 \\
$\Move$ & \{forward, back, left, right, 4 diagonals, idle\} & 9 \\
$\Dodge$ & \{dodge, idle\} & 2 \\
$\HA$ & \{light attack, heal (consume flask), idle\} & 3 \\
\bottomrule
\end{tabular}
\end{table}

\paragraph{Skill-Specific Observations and Actions.}
The exact observation vectors and actions used in code are summarized in Tables~\ref{tab:obs_spaces} and~\ref{tab:action_spaces}. All observation variables not explicitly stated as normalized are treated as raw floats/integers.

\subsection{Skill-Specific Reward Objectives}
\label{app:reward}

Each skill is trained with its own reward $r_t^k$ that targets a narrow responsibility and is computed from the same process-memory interface used for observations.
The shaping terms depend only on generic combat variables (e.g., angles, distances, HP/stamina ratios, lock status, remaining flasks, and terminal events) and do not encode boss-specific action sequences or phase-specific scripts.
Weights are chosen so that per-episode returns fall in the scales reported in Section~\ref{sec:experiments}, improving stability across skills.

\paragraph{Camera ($\Cam$).}
Camera control is rewarded for reducing the camera--target angle.
A dense term is used
\begin{equation}
r_t^{\Cam} \;=\; -\alpha_t,
\end{equation}
optionally combined with a small bonus $\mathbb{I}[\alpha_t < \varepsilon_{\Cam}]$ when alignment is within a tolerance to encourage maintaining lockable framing. In our experiments $\varepsilon_{\Cam} = 0.6$ was used.

\paragraph{Lock-on ($\Lock$).}
Lock-on is rewarded for maintaining a valid lock state while the target is in view
\begin{equation}
r_t^{\Lock} \;=\; \mathbb{I}[\ell_t=1]\;-\;\mathbb{I}[\ell_t=0],
\end{equation}
with lock toggling treated as a discrete action and encouraged only when it improves sustained lock status.

\paragraph{Movement ($\Move$).}
Movement is rewarded for reaching and maintaining engagement configurations rather than wandering.
A distance-based term is used
\begin{equation}
r_t^{\Move} \;=\; -\lambda_\Move d_t,
\end{equation}
where $\lambda_\Move = 1/10$ in our experiments. Optionally, a term $\mathbb{I}[d_t < d_{\min}\, \wedge\, \texttt{Sideways Movement}]$ can be used to encourage circling around the enemy when close by. Surprisingly, the agent learns to do so even without the optional part in a few more training timesteps.

\paragraph{Dodge ($\Dodge$).}
Dodging is rewarded for survival and damage avoidance, while discouraging wasteful defensive actions.
Let $\Delta h_t = h_t-h_{t-1}$ be the change in player HP ratio (negative under damage).
A typical term is
\begin{equation}
r_t^{\Dodge} \;=\; \lambda_{\text{alive}} \;+\; \lambda_{\text{dmg}} \Delta h_t\;-\;\lambda_{\text{dead}}\mathbb{I}[\texttt{Agent Dead}]\;-\;\mathbb{I}[\sigma_t < \varepsilon_\Dodge],
\end{equation}
where in our experiments $(\lambda_{\text{alive}}, \lambda_{\text{dmg}}, \lambda_{\text{dead}}, \varepsilon_\Dodge) = (0.02, 5, 5, 0.05)$.
The terminal penalty is applied on death. Optionally instead of $\mathbb{I}[\sigma_t < \varepsilon_\Dodge]$ a more dense reward defined by $\Delta \sigma_t = \sigma_t-\sigma_{t-1}$ can be used to continuously signal stamina consumption.
This construction makes poorly timed dodges harmful (via the induced damage term) and yields a slowly improving return curve when survival time increases.

\paragraph{Heal--attack ($\HA$).}
The heal--attack policy is rewarded for trading off damage dealt, damage taken, and resource usage.
Let $\Delta h^e_t = h^e_t-h^e_{t-1}$ denote enemy HP change (negative when damage is dealt), and let $\Delta n^{\text{estus}}_t$ denote the change in remaining flasks (negative when consumed).
The objective is
\begin{equation}
r_t^{\HA} \;=\; \lambda_{\text{dmg}} \Delta h_t\;-\;\lambda_{\text{hit}} \Delta h^e_t\;-\;\lambda_{\text{dead}}\mathbb{I}[\texttt{Agent Dead}]\;+\;\lambda_{\text{success}}\mathbb{I}[\texttt{Enemy Dead}], \label{eq:ha_reward}
\end{equation}
where in our experiments we used $(\lambda_{\text{dmg}}, \lambda_{\text{hit}}, \lambda_{\text{dead}}, \lambda_{\text{success}}) = (5, 15, 5, 5)$. To penalize resource consumption we can use the same method as in $\Dodge$. Either of $\mathbb{I}[\sigma_t < \varepsilon_\HA]$ or $\Delta \sigma_t$ can be used to penalize overcommitting. Note that healing is rewarded using the term $\Delta h_t$ which encourages healing when health is low.
Because the number of flasks per episode is capped (one in Phase~1 and two in Phase~2), transitions containing heal events remain sparse, which makes credit assignment for heal timing significantly harder than learning an attack policy.

\section{End-to-End Baseline}
\label{app:end2endbase}

To contextualize the benefits of the directed skill-graph factorization, we train an atomic end-to-end baseline: a single DQN policy that directly controls camera, lock-on, locomotion, dodging, and attack/heal decisions through one discrete action space. The baseline uses the same process-memory state interface and training setup as the modular agent. The only difference is that control is learned with a single monolithic policy.

The end-to-end agent observes the full global state $s_t$ (25 dimensions), i.e., the union of the features used across the skill-specific policies. Concretely, we use
\begin{align}
o_t^{\text{e2e}} = s_t = [\kappa_t^{e},\ \rho_t^{e},\ \kappa_t^{p},\ \rho_t^{p},\ \sigma_t,\ h_t,\ h_t^{e},\ O_t^{e},\ O_t^{p},\ n_t^{\text{estus}},\ d_t,\ \hat{\vd}_t^{\top},\ \vc_t^{\top},
\vp_t^{\top},\ \ve_t^{\top},\ \alpha_t,\ \ell_t].
\end{align}

The baseline uses a 16-action discrete set that merges the skill action subspaces:
(i) movement (forward/back/left/right),
(ii) movement+dodge in each cardinal direction,
(iii) camera adjustments (up/down/left/right),
(iv) toggle lock-on,
(v) light attack,
(vi) heal (consume flask),
and (vii) idle.

To avoid misleading the end-to-end policy with a mixture of skill-specific shaping signals, we train the end-to-end baseline using the same reward as the heal--attack objective (Eq.~\ref{eq:ha_reward}). This keeps the supervision consistent with the overall combat objective and avoids injecting additional dense signals that are specific to individual skills.

\begin{figure}[t!]
  \centering
  \begin{subfigure}[b]{0.49\linewidth}
    \centering
    \includegraphics[width=\linewidth]{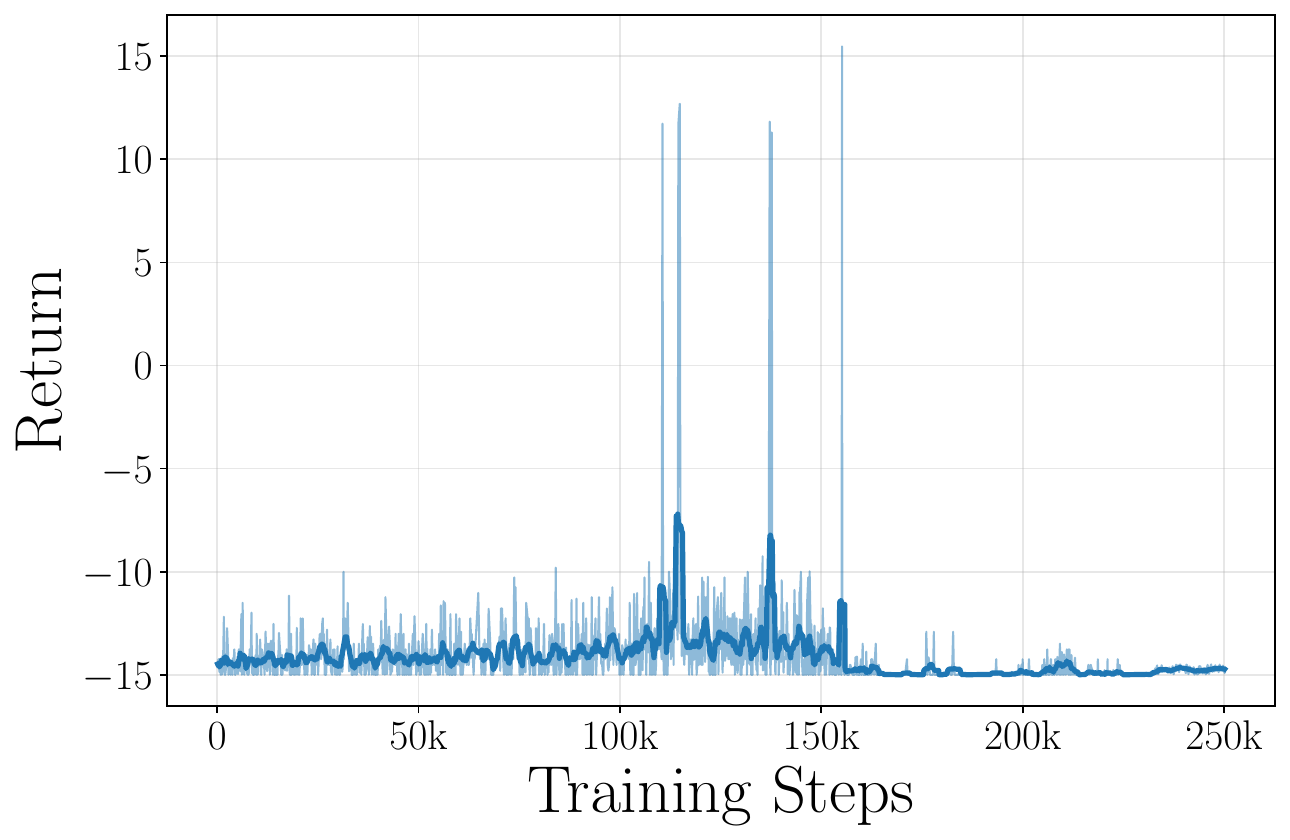}
    \caption{Training return.}
    \label{fig:e2e_return}
  \end{subfigure}\hfill
  \begin{subfigure}[b]{0.49\linewidth}
    \centering
    \includegraphics[width=\linewidth]{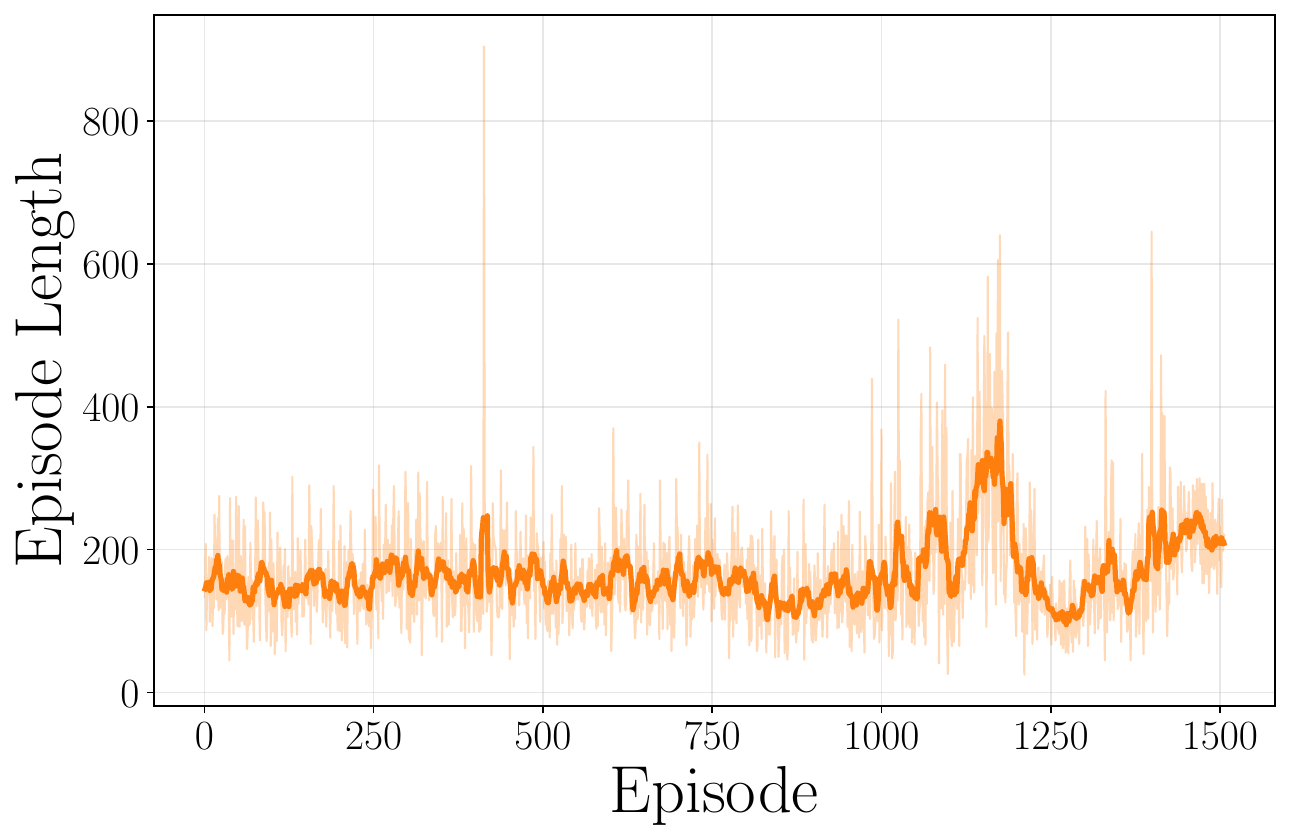}
    \caption{Episode length.}
    \label{fig:e2e_length}
  \end{subfigure}
  \caption{End-to-end baseline struggles under the same state interface.
  The DQN baseline quickly plateaus, exhibiting little change in return or survival time beyond early training. The episode length plot shows that the agent couldn't even postpone early deaths even with its conservative strategy. The translucent line shows the raw value, and the darker line shows the 10-period moving average.}
  \label{fig:e2e_training}
\end{figure}

Figure~\ref{fig:e2e_training} shows the training curves for return and episode length. Although we initially planned to train for 500k steps, the baseline plateaued early (already by $\sim$250k steps) with no meaningful improvement, so we stopped training due to the high wall-clock cost. Qualitatively, the learned behavior collapses into a poor survival heuristic: the agent typically locks on and repeatedly dodges backward to avoid immediate death, without learning a reliable dodging policy (timing/direction) or an effective attack strategy.

\section{Experimental Details}
\label{app:experimental_details}

\paragraph{Implementation and Algorithm.}
All skills are trained with Deep Q-Networks (DQN) using Stable-Baselines3.
Unless otherwise noted, we use the default SB3 DQN hyperparameters and a single environment instance (no parallel rollouts).
Observations are normalized using \texttt{VecNormalize} and rewards are not normalized.

We use skill-specific maximum episode lengths to match the temporal scope of each subtask:
128 steps for $\Cam$ and $\Move$, 64 for $\Lock$, 512 for $\Dodge$,  1024 for $\HA$, and 2048 for end-to-end baseline. For $\Dodge$, $\HA$, and end-to-end agents the episode also terminates upon death or success.
Episodes also terminate on standard environment termination conditions (e.g., agent death or boss defeat).

\paragraph{Hyperparameters.}
Across all skills, we use a learning rate of $3\times 10^{-4}$ and batch size of 256.
For the modular skills, replay buffer capacity is set to approximately one-third of the corresponding training interaction budget.
For the end-to-end baseline, we use a replay buffer size of 100{,}000 transitions.
Other optimizer and DQN parameters follow the SB3 defaults (including the exploration schedule and target network update settings).

\paragraph{Initialization and State Randomization.}
For the upstream skills ($\Cam$, $\Lock$, $\Move$), episodes are initialized by placing the player at random locations within the arena to diversify camera/targeting/locomotion configurations and avoid overfitting to a single spawn state.

To avoid full game restarts during training, in-game death handling is disabled.
Instead, we treat the agent as ``dead'' and terminate the episode when the normalized player HP falls below $h_t < 0.05$.
Episodes also terminate on boss defeat and are truncated at the skill-specific horizon limits reported above.

\paragraph{Action Execution Timing.}
Because the environment is controlled through key-press macros, each discrete action is implemented with a fixed real-time delay to allow the in-game animation/transition to register before the next observation.
For $\Cam$, $\Lock$, and $\Move$, each action step uses a $0.1$s delay.
For $\Dodge$, a dodge action occupies $0.5$s.
In $\HA$, light attack uses $0.2$s, drinking Estus uses $0.3$s, and attempting to drink with zero flasks uses $0.5$s (a no-op delay).
The idle action uses $0.1$s.

\section{Additional Results}
\label{app:additional_results}

\subsection{$\Dodge$ Analysis}
\label{app:dodge_analysis}

For skills whose reward is strongly correlated with survival (notably $\Dodge$), episode length provides an interpretable diagnostic: longer episodes typically indicate fewer early deaths and improved defensive timing. We report episode length curves to make learning progress easier to interpret.

\begin{figure}[t]
  \centering
  \includegraphics[width=0.65\linewidth]{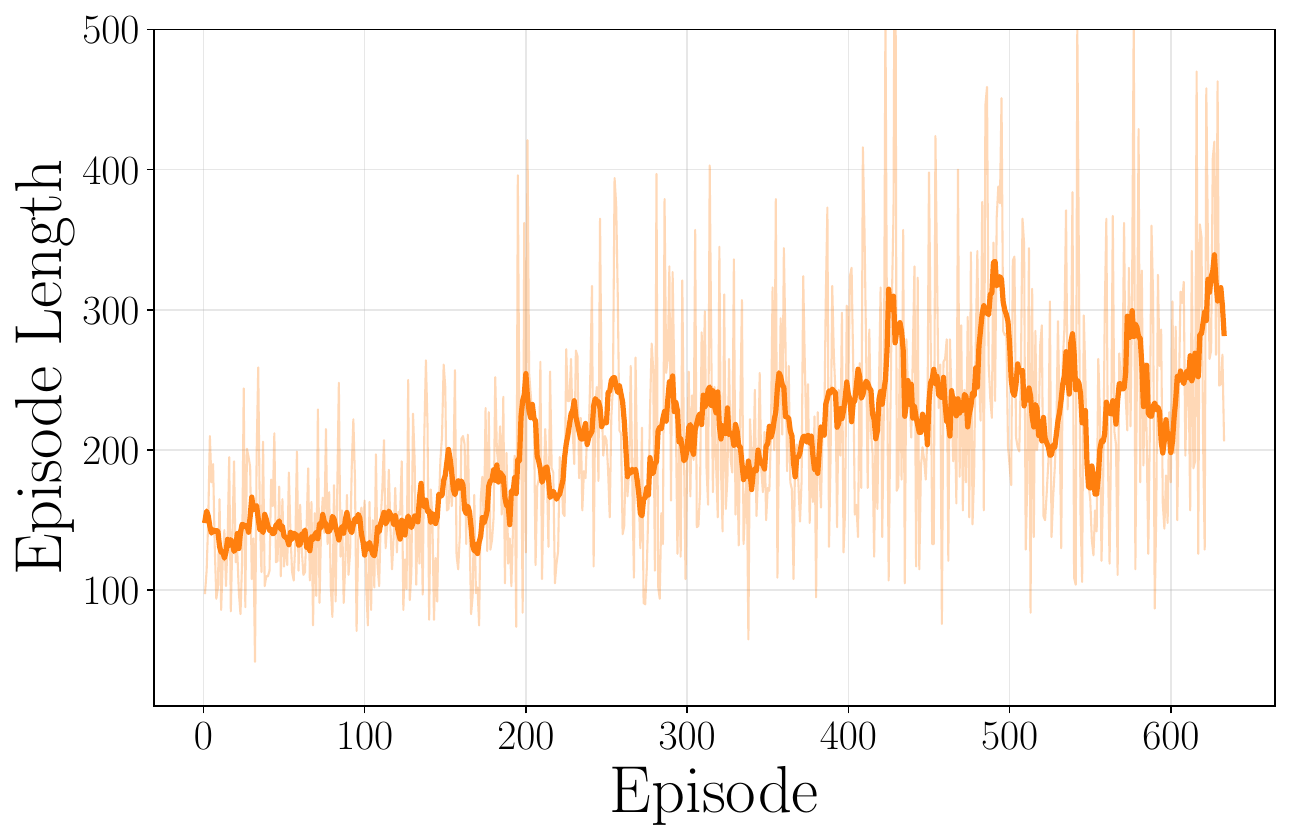}
  \caption{Episode length of $\Dodge$ agent vs. training episodes.}
  \label{fig:dodge_ep_len}
\end{figure}

 Figure~\ref{fig:dodge_ep_len} shows that the mean episode length increases from roughly $150$ to $300$ steps over training, indicating that even without access to healing (no Estus use during $\Dodge$ training), the learned policy survives about $2\times$ longer than a random-dodge baseline, consistent with a competent defensive timing strategy.

\end{document}